\documentclass{article}

\PassOptionsToPackage{authoryear,square, sort}{natbib}
\usepackage[preprint]{neurips_2024}

\usepackage[utf8]{inputenc}
\usepackage[T1]{fontenc}
\usepackage{lmodern}
\usepackage{amsmath,amsfonts, amssymb}
\usepackage{booktabs}
\usepackage{graphicx}
\usepackage{enumitem}   
\usepackage{xcolor}
\usepackage{url}
\usepackage{colortbl}
\usepackage{tcolorbox}
\usepackage{booktabs}
\usepackage{float}
\usepackage{algorithm,algorithmicx,algpseudocode}
\tcbuselibrary{breakable, skins, listings}

\tcbset{
  uibox/.style={
    enhanced, breakable, frame hidden, colback=white,
    boxsep=6pt,left=8pt,right=8pt, before skip=4pt, after skip=4pt,
    overlay middle={
      \draw[hotpink!70!black,line width=1.5pt]
        (frame.north west)--(frame.south west);
      \draw[hotpink!70!black,line width=1.5pt]
        (frame.north east)--(frame.south east);
    },
    overlay first={
      \draw[hotpink!70!black,line width=1.5pt]
        ([yshift=-3pt]frame.north west)--(frame.south west);
      \draw[hotpink!70!black,line width=1.5pt]
        ([yshift=-3pt]frame.north east)--(frame.south east);
      \draw[hotpink!70!black,line width=1.5pt]
        ([xshift=3pt]frame.north west)--([xshift=-3pt]frame.north east);
      \draw[hotpink!70!black,line width=1.5pt,rounded corners=3pt]
        ([yshift=-3pt]frame.north west)--(frame.north west)--([xshift=3pt]frame.north west);
      \draw[hotpink!70!black,line width=1.5pt,rounded corners=3pt]
        ([yshift=-3pt]frame.north east)--(frame.north east)--([xshift=-3pt]frame.north east);
    },
    overlay last={
      \draw[hotpink!70!black,line width=1.5pt]
        (frame.north west)--([yshift=3pt]frame.south west);
      \draw[hotpink!70!black,line width=1.5pt]
        (frame.north east)--([yshift=3pt]frame.south east);
      \draw[hotpink!70!black,line width=1.5pt]
        ([xshift=3pt]frame.south west)--([xshift=-3pt]frame.south east);
      \draw[hotpink!70!black,line width=1.5pt,rounded corners=3pt]
        ([xshift=3pt]frame.south west)--(frame.south west)--([yshift=3pt]frame.south west);
      \draw[hotpink!70!black,line width=1.5pt,rounded corners=3pt]
        ([xshift=-3pt]frame.south east)--(frame.south east)--([yshift=3pt]frame.south east);
    },
  }
}
\definecolor{hotpink}{RGB}{255,105,180}
\usepackage{array}  
\usepackage{caption} 
\usepackage{hyperref}   

\title{UI-Bench: A Benchmark for Evaluating Design Capabilities of AI Text-to-App Tools}

\usepackage{authblk}    
\usepackage{hyperref}   

\author{\vspace{-0.5em}\textbf{
  Sam Jung$^{\heartsuit\diamondsuit}$ \quad
  Agustin Garcinuno$^\diamondsuit$ \quad
  Spencer Mateega$^\diamondsuit$
}}
\affil{
  $^\heartsuit$University of Pennsylvania \quad
  $^\diamondsuit$AfterQuery
}
\affil{\vspace{-0.5em}\texttt{\{samj,agustin,spencer\}@afterquery.com}

\url{https://uibench.ai/leaderboard}}
\affil{}

\begin{document}
\maketitle

\begin{abstract}
AI text-to-app tools promise high quality applications and websites in minutes, yet no public benchmark rigorously verifies those claims.
We introduce \textsc{UI-Bench}, the first large-scale benchmark that evaluates visual excellence across competing AI text-to-app tools through expert pairwise comparison. Spanning \textit{10} tools, \textit{30} prompts, \textit{300} generated sites, and \textit{4000+} expert judgments, \textsc{UI-Bench} ranks systems with a TrueSkill-derived model that yields calibrated confidence intervals. \textsc{UI-Bench} establishes a reproducible standard for advancing AI-driven web design. We release (i) the complete prompt set, (ii) an open-source evaluation framework, and (iii) a public leaderboard. The generated sites rated by participants will be released soon. 
\end{abstract}

\section{Introduction}
\subsection{Rapid Emergence of of AI App and Website Tools}
Low-code and no-code development platforms have emerged as powerful prototyping tools, allowing organizations to rapidly create applications with minimal coding effort. This wave has accelerated the market for AI text-to-app tools.
TechRadar's 2025 survey alone
``tested, rated, and ranked'' the \textit{nine} best AI tools out of more than eighty tools reviewed
\citep{TechRadar2025}.  While marketing copy promises “professional quality in one click,” most discourse remains anecdotal, consisting mostly of feature lists and speed claims. They offer little systematic evidence about the visual quality these tools actually produce.

\subsection{Design Quality Matters}
Visual excellence directly impacts business outcomes.  The seminal Stanford Web Credibility
study found that 46{\%} user comments cited a site's \emph{look and feel} when judging
credibility \citep{Fogg2003}, and other studies support this fact as well \citep{Robins2008}. Empirical UX studies show that both high visual intensity (e.g., animation, flashing, cluttered interfaces) and excessive content volume impair user engagement; visual overload triggers strong negative reactions before conversions climb, and simpler, content-light landing pages consistently convert better than dense, information-heavy versions 
\citep{Jankowski2019,DvirGafni2018}. For AI tools to succeed in market adoption, their output must excel on these perception-critical dimensions.

\subsection{The Evaluation Challenge}
\label{evalchallenge}
Aesthetic quality lacks a single objective ground truth, and is fundamentally perception-driven \citep{LavieTractinsky2004,Moshagen2010,Ngo2003}. Previous HCI work reports only moderate agreement across raters and substantial unexplained variance in aesthetic evaluations \citep{Wangenheim2018,Reinecke2013}. Public, non\mbox{-}blind crowd voting also suffers from social\mbox{-}influence herding: a single early up\mbox{-}vote can raise a post’s final score by about 25\% \citep{muchnik2013}. 

For this study, we use blinded pairwise judgements from expert designers as the primary endpoint. We omit automated image proxies as primary metrics because their alignment with human aesthetic preference is weak in our setting. For instance, though FID and CLIP are widely used automated proxies for image quality and alignment~\citep{heusel2017fid,radford2021clip}, recent work shows such proxies can mis-rank or even contradict human preferences—especially for aesthetics and design-heavy images~\citep{jayasumana2024rethinkingfid,Kirstain2023PickaPic,Xu2023ImageReward,Lee2023HEIM}; see~\S\ref{automated-aesthetic} for a survey and scope caveats. Procedural details for our pairwise protocol and ranking appear in~\S\ref{methods}.

\subsection{Contributions}
\begin{enumerate}[leftmargin=*,nosep]
  \item \texttt{\textbf{UI-Bench}}: a fully reproducible protocol for assessing visual design quality of AI text-to-app tools across realistic prompt categories.
  \item \textbf{Statistical Framework}: an adapted TrueSkill model that delivers stable rankings with 95{\%} confidence intervals from sparse pairwise votes.
  \item \textbf{Public Resources}: a public leaderboard.
\end{enumerate}
\textsc{UI-Bench} closes the gap between feature-checklist reviews and academic usability tests by supplying the first benchmark that directly measures professional-grade visual design quality in AI text-to-app tools.

\section{Related Work}

\subsection{From Ground-Truth to Preference Benchmarks}
\textbf{LLM Benchmarks.} Static, ground-truth benchmarks such as
\textsc{MMLU}~\citep{hendrycks2020mmlu}, \textsc{HellaSwag}~\citep{zellers2019hellaswag},
\textsc{GSM-8K}~\citep{cobbe2021gsm8k}, \textsc{FinanceQA}~\citep{mateega2025}, \textsc{AGIEval}~\citep{zhong2023agieval},
\textsc{BIG-Bench}~\citep{srivastava2023bigbench}, and
\textsc{HumanEval}~\citep{chen2021humaneval} dominate language-model
evaluation.
Alongside static benchmarks, recent work has relied on human or LLM-judge based pairwise evaluations: \textsc{MT-Bench}~\citep{zheng2023mtbench} and \textsc{AlpacaEval}
\citep{li2023alpacaeval} use LLMs as judges on curated prompts, while \textsc{Chatbot Arena}~\citep{Chiang2024Arena} aggregates crowd pairwise votes with an Bradley-Terry-style rating for leaderboards.

\subsection{Existing Web Development Benchmarks}
Most recent work targets design$\rightarrow$code or webpage$\rightarrow$code generation and leans on automatic checks rather than blinded expert preference on finished designs. \textsc{Web2Code} releases a large webpage$\rightarrow$code dataset and evaluation suite with rendered-image comparison and GPT\mbox{-}4V scoring \citep{yun2024web2code}. \textsc{WebCode2M} provides 2.56M real design–code pairs and establishes a benchmark for training and evaluating models in automated webpage code generation, prioritizing scalability over aesthetic quality \citep{gui2025webcode2m}.
 \textsc{DesignBench} frames front-end generation/edit/repair across React/Vue/Angular with programmatic evaluation \citep{xiao2025designbench}. \textsc{FrontendBench} automates sandboxed tests and reports 90.45\% agreement with experts on specific UI behaviors, not holistic taste \citep{zhu2025frontendbench}. \textsc{WebGen\mbox{-}Bench} is the closest task analogue: it generates full sites from instructions, verifies functionality via a navigation agent, and grades appearance with GPT\mbox{-}4o (1–5); it does not use blinded expert pairwise judgments \citep{lu2025webgenbench}. In terms of human validation, \textsc{Design2Code} is closest. It curates 484 webpages and validates visual-similarity metrics against human ratings, but it benchmarks multimodal language model (MLLM) capabilities by reproducing webpages from original screenshots rather than by comparing preferences between competing tools \citep{si2025design2code}. Though existing benchmarks emphasize functionality, fidelity, and code similarity; none adopt blinded expert pairwise preference to evaluate visual design quality, and none evaluate AI text-to-app tools.

\subsection{Automated Aesthetic Metrics and GUI Heuristics}
\label{automated-aesthetic}
Generative-image quality is typically scored by Inception Score (IS)~\citep{salimans2016is}, Fr\'echet Inception Distance (FID)~\citep{heusel2017fid}, or CLIP-based similarity~\citep{hessel2021clipscore}. Classical GUI--layout heuristics compute rule-based scores for symmetry, balance, and rhythm~\citep{Ngo2003}; later work trains neural predictors on curated photography and poster datasets~\citep{lu2014dpchallenge,odonovan2014color}.

For multi-section webpages (rather than single-image quality), automated aesthetic scorers exhibit limited cross-dataset generalization. Domain-tailored predictors for webpage aesthetics can achieve promising \emph{in-domain} correlations (e.g., Webthetics reports $r \approx 0.85$)~\citep{dou2019webthetics}; however, correlation on a single dataset does not guarantee faithful \emph{pairwise} rankings or robustness across datasets. Similarly, generic proxies such as FID and CLIP-based scores—while widely used~\citep{heusel2017fid,radford2021clip, hessel2021clipscore}—have been shown to misalign with, and sometimes contradict, human preferences for aesthetics and design-heavy images~\citep{jayasumana2024rethinkingfid,Kirstain2023PickaPic,Xu2023ImageReward,Lee2023HEIM}. We therefore treat automated scores as ancillary diagnostics, and select blinded expert pairwise judgement as our primary endpoint (see~\S\ref{evalchallenge}).

\subsection{Gap}
None of the above benchmarks evaluate the holistic UI and UX quality of AI tools,
leaving researchers and practitioners without a standard ranking metric.

\section{Methodology}
\label{methods}

\subsection{Theoretical Framework}

\subsubsection{Why Pairwise Comparison}
We chose pairwise comparison because absolute ratings are noisy: scale interpretation and anchoring make
designers disagree. Thurstone’s comparative-judgment theory offers a clean alternative to scaled methods, with previous deployments of adaptive comparative judgment reporting very high rank-order reliability ($\approx$0.95–0.97) \citep{pollitt2012cj,whitehouse2012acj}.
Early psychometrics formalised preference as \emph{comparative judgment}
\citep{thurstone1927law}; the Bradley--Terry model (BT) \citep{bradley1952rank}
and Bayesian extensions such as TrueSkill \citep{herbrich2006trueskill}
remain standard for converting pairwise outcomes into system skill estimates.
This method also scales: Chatbot Arena logged 240k+ human comparisons and produced
a stable BT leaderboard for 50+ LLMs \citep{Chiang2024Arena}, with
formal convergence analysis reported in that work.
\textsc{UI-Bench} therefore collects only binary designer preferences and fits a
TrueSkill model; implementation details are provided in~\S\ref{sec:tourney}.

\subsubsection*{Defining Design Quality}\label{sec:def_quality}
We define design quality as the probability that a professional designer prefers layout A over layout B on the client-delivery question:
\begin{quote}
\small
\textit{“Which project would you be more likely to deliver to a client?”}
\end{quote}

This single forced choice captures holistic judgement across UI and UX completeness without a rigid rubric. Because professional designers are trained evaluators, aggregating their independent judgements via expert panels is a validated approach for ranking creative work \citep{amabile1982cat,amabile1996cat}. We use expert comparative judgement: designers make binary pairwise choices on the client-delivery question. This method is established in educational assessment with high reliability \citep{pollitt2012cj}, and the same pairwise-preference framing also underlies RLHF work \citep{christiano2017rlhf}.

\paragraph{Evaluation Framing.}
A generic “Which one do you like more?” question risks collapsing professional standards 
into personal taste. To ensure judgments are tied to design practice, \textsc{UI-Bench} frames 
comparisons as a client-delivery question. This anchors evaluations on holistic UI and UX quality within a professional context.

\paragraph{Design of the Evaluation Protocol.}
Our protocol incorporates several design decisions that address known shortcomings:
\begin{itemize}
  \item \textbf{Full-view requirement.} Raters must view the entire webpage in fullscreen mode before 
  voting, ensuring assessments reflect the complete composition rather than a cropped 
  hero or single section.
  \item \textbf{Prompt standardization.} \textsc{UI-Bench} uses synthetically generated prompts across a range of situations and tasks so that 
  outcomes reflect tool capability, not rater creativity in crafting prompts. See ~\S\ref{promptdesignandgeneration} for prompt generation protocol.
  \item \textbf{Blinding and quality controls.} Randomized left–right placement, hidden 
  tool identities, and split voting sessions further reduce bias and noise relative to open 
  crowd arenas.
\end{itemize}
These design decisions incrementally add rigor beyond prior arena-style studies and 
establish \textsc{UI-Bench} as a reproducible standard for evaluating the visual design quality of AI 
text-to-app tools.

\subsection{Tournament Structure}\label{sec:tourney}

\paragraph{Randomised Pairwise Arena.}
We run a continuous arena where each vote compares two tools on one prompt for a single expert. Prompt selection has two phases: a base-coverage phase that shows each expert all 30 prompts exactly once (by round 30), followed by an adaptive phase that samples from the 30 prompts based on a score calculated by recency, uncertainty, and the score gap between tools. We draw uniformly from this top-5. Pair selection is not uniform: we score every candidate pair by a weighted blend of exposure balance, opponent novelty, uncertainty $(\sigma_a+\sigma_b)$, and TrueSkill match quality, then apply penalties for pairs already seen by the expert, recent cooldown, repeat-cap exceedance, and hot-tool over-exposure. We pick the highest-scoring pair with deterministic tie-breakers (lower pair count, then lower max exposure). Tools are rated per prompt; the headline score for each tool is the mean TrueSkill $\mu$ across the 30 prompts.

\paragraph{TrueSkill Rating System.}\label{sec:trueskill}
For each prompt, every tool starts at $\mu=25.0$ and $\sigma=8.33$. After each 1-v-1 outcome, we update the two prompt-specific ratings within a TrueSkill environment (no drift, no draws) \citep{herbrich2006trueskill}. The global tool score is the mean of prompt-level $\mu$ across the 30 prompts.

\subsection{Expert Evaluators}\label{sec:experts}

\paragraph{Inclusion Criteria.}
We hand selected and invited the experts in this study on the basis of having professional UI and UX experience as a designer, web developer, researcher, or other relevant profession.

\paragraph{Demographics (N=\textit{194}).}
Our panel spans multiple roles (multi-select; totals exceed 100\%): Designers~(\textit{65.5\%}, 127), Web Developers~(\textit{57.2\%}, 111), Researchers~(\textit{60.3\%}, 117), Other~(\textit{7.2\%}, 14). When asked whether they had previously used AI tools for creating applications or websites, \textit{10.8\%} (21) answered yes and \textit{89.2\%} (173) answered no.

\section{Experimental Setup}

\subsection{Prompt Design and Generation}
\label{promptdesignandgeneration}

\paragraph{Taxonomy and Counts.}
We generated 30 prompts across five categories (6 per category): 
Marketing/Landing, Editorial/Blog, Portfolio/Case Study, E-commerce, and Local/Service. 
Within each category we include 4 \emph{websites} and 2 \emph{webapps}. 
Websites are read-mostly marketing or informational pages; simple forms are permitted, 
but no authentication or persistent data storage is included. 
Webapps are stateful experiences with saved data and at least two screens 
(e.g., list/table plus create/edit form), incorporating basic validation and 
empty/loading/error states. Authentication flows were excluded to avoid confounding 
design evaluation with account-management UX.

\paragraph{Authoring Model.}
Prompts were synthetically generated with GPT-5 (Thinking) using ChatGPT on August 20, 2025. We report the model, interface, and date to document the generation 
conditions.

\paragraph{Prompt Specification (summary).}
We instructed the model to: 
(i) write realistic client briefs of 150–180 words; 
(ii) specify a primary goal/CTA (lead, purchase, booking, subscribe, donate, apply, RSVP, informational); 
(iii) include a one‑sentence brand vibe (1–2 adjectives); 
(iv) optionally add non‑visual constraints (e.g., “include pricing tiers” or “include a contact form”); 
(v) avoid layout directives and authentication/login requirements. 
The exact authoring prompt is provided in Appendix~\ref{app:authoring-prompt}.

\paragraph{Validation and Filtering.}
We applied light editorial review, requiring each prompt to be at least 150 words, maintain diverse category coverage with correct counts, and include a coherent CTA. Prompts that failed these checks were regenerated.

\paragraph{Release.}
We release the final prompts (title, type, sector, goal, scenario, vibe, constraints) in Appendix~\ref{app:toolprompts}. Future versions of this benchmark will include a scripted generation pipeline.

\subsection{Tools Evaluated}

We evaluated ten anonymized AI text-to-app tools, representing a mix of leading commercial platforms: 
v0 by Vercel, Bolt, Orchids, Lovable, Replit, Figma Make, Base44 by Wix, Magic Patterns, Same.new, and Create Anything. 
For consistency, all generations were produced between August~20th--22,~2025. 
Each tool was run with its default configuration settings, without manual intervention or prompt engineering.

\subsection{UI Generation Protocol}
\subsubsection{Controlled Generation}
The authors generated all designs with a
60-minute time limit per prompt per tool. Each tool received up to two independent one-shot attempts per prompt. For each fresh attempt, we permitted up to two repair passes to fix blocking errors that prevented rendering. We stopped at the first usable output (the full page displaying). This lenient protocol ensured every tool produced at least one functional UI. Some tools (e.g., Same.new) intermittently stalled during generation, typically on file-read steps, consistent with brittle agent orchestration or system prompts.

\subsubsection{Output standardization}
Websites are displayed in iFrames, with sanitized URLs to prevent tool identification.

\subsection{Evaluation Interface}
\subsubsection{Onboarding}
Experts received a special code to log in to the study, ensuring only vetted participants could participate. We collected the following from experts before allowing them to vote: first name, last name, profession (Designer, Web Developer, Researcher, or Other), and whether they have previously used an AI website generating tool before.
\subsubsection{Comparison Presentation}
\begin{figure}[t]  
  \centering
  \includegraphics[width=\linewidth]{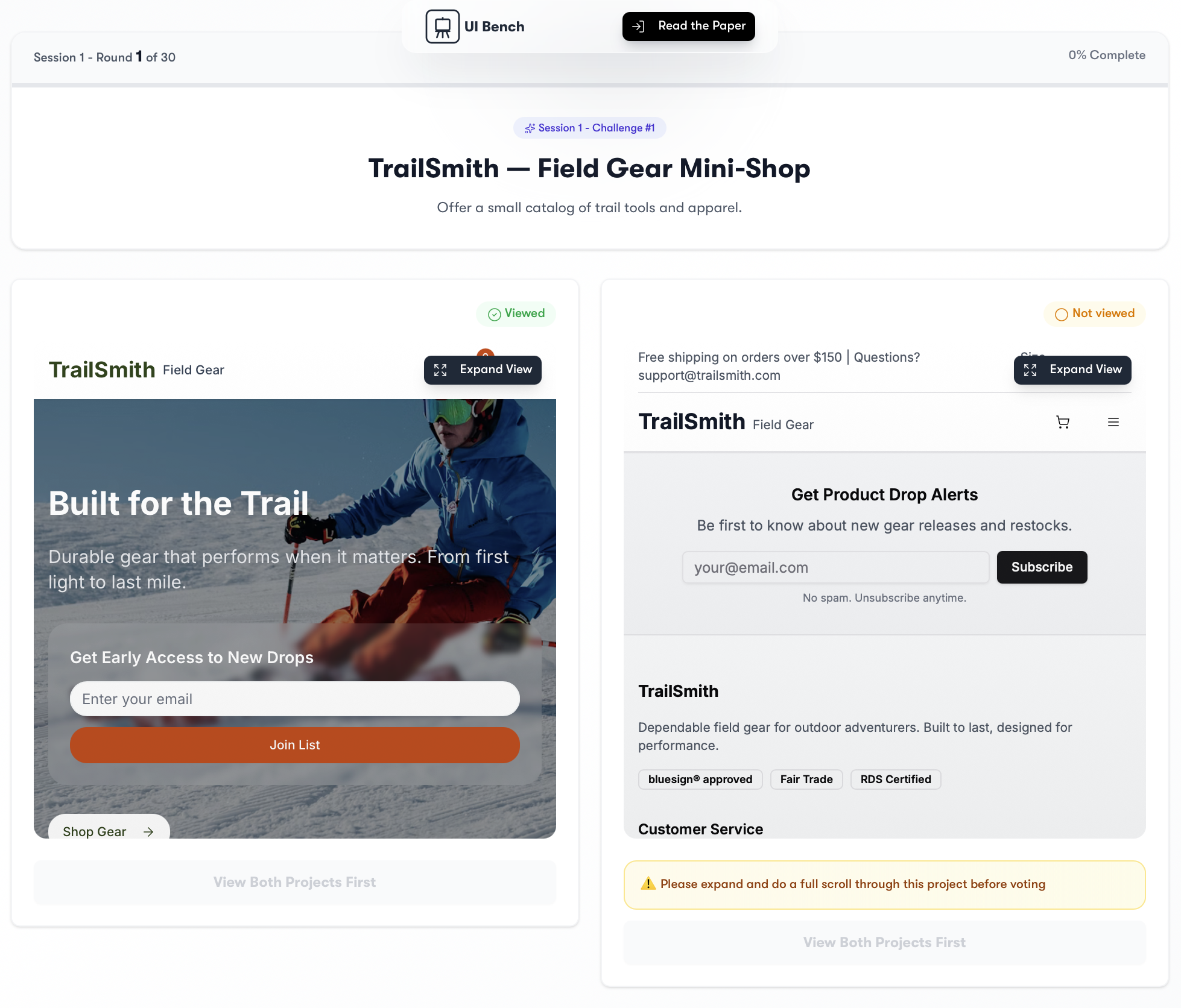}
  \caption{Example comparison interface used in \textsc{UI-Bench}. Raters view two webpages in a blinded layout with randomized placement and a full-view requirement.}
  \label{fig:comparison-ui}
\end{figure}
\noindent Each comparison is presented in a controlled, blinded UI. As shown in Figure~\ref{fig:comparison-ui}, we required (i) a full-view requirement, (ii) incorporate split voting sessions, (iii) hide tool identities, (iv) standardize prompts, and (v) randomize left-right placement.

\subsection{The Question}
For each pair of designs generated for a prompt, raters answer a single forced-choice question as described in~\S\ref{sec:def_quality}. The following instructions are shown to experts before they started each voting session: 

\begin{tcolorbox}[colframe=hotpink!70!black, colback=white, boxrule=1.5pt, arc=3pt, outer arc=3pt, breakable]
\begin{verbatim}
In this study, you will see two projects side by side, one on the left 
and one on the right. For each pair, vote according to this metric: 
"Which project would you be more likely to deliver to a client?"
\end{verbatim}
\end{tcolorbox}

\noindent Responses were binary: \emph{"Vote for this Project"} was presented under both left and right options. No tie option was provided.

\subsubsection{Quality Controls}
Each expert was instructed to vote on 60 matches, administered in two sessions of 30 votes each, with an optional third bonus session for up to 90 total votes. Breaks were permitted between sessions, and a one-day interval between sessions was recommended to maintain attention. As participation was voluntary, the 60-vote target served as guidance rather than a strict requirement.

\section{Results}

\subsection{Leaderboard Rankings}

Table~\ref{tab:leaderboard} reports the overall leaderboard aggregated across 
$n=4{,}075$ blinded pairwise matches, and $n=8{,}150$ unique tool comparisons. Ratings are reported as posterior means 
($\mu$) under the TrueSkill model, alongside posterior uncertainty ($\sigma$), 
95\% credible intervals, and empirical win rates.

\begin{table}[h]
  \centering
  \caption{Overall Leaderboard Results ($n=4{,}075$ unique matches)}
  \label{tab:leaderboard}
  \begin{tabular}{r l c c c}
    \toprule
    Rank & Tool & Rating ($\mu$) & Uncertainty + 95\,\% CI & Win Rate \\
    \midrule
    1 & Orchids         & 30.12 & $\sigma=1.77$ \; [26.61, 33.55] & 67.5\,\% \\
    2 & Figma Make      & 27.46 & $\sigma=1.71$ \; [24.11, 30.81] & 57.1\,\% \\
    3 & Lovable         & 27.14 & $\sigma=1.72$ \; [23.77, 30.51] & 54.8\,\% \\
    4 & Anything        & 25.46 & $\sigma=1.69$ \; [22.15, 28.77] & 51.2\,\% \\
    5 & Bolt            & 24.44 & $\sigma=1.68$ \; [21.15, 27.73] & 48.9\,\% \\
    6 & Magic Patterns  & 24.23 & $\sigma=1.70$ \; [20.90, 27.56] & 47.0\,\% \\
    7 & Same.new        & 23.57 & $\sigma=1.70$ \; [20.24, 26.90] & 45.8\,\% \\
    8 & Base44 by Wix   & 23.47 & $\sigma=1.69$ \; [20.16, 26.78] & 47.4\,\% \\
    9 & v0              & 22.24 & $\sigma=1.72$ \; [18.87, 25.61] & 41.2\,\% \\
    10 & Replit         & 20.95 & $\sigma=1.73$ \; [17.56, 24.34] & 38.9\,\% \\
    \bottomrule
  \end{tabular}
\end{table}

Orchids leads with a mean rating of 30.12 and a 67.5\% win rate, 
substantially ahead of the second-ranked system (Figma Make, $\mu$=27.46, 
57.1\%). Lovable rounds out the top three with $\mu$=27.14 and a 54.8\% win 
rate. Mid-tier tools cluster closely: Anything, Bolt, and Magic Patterns 
all exhibit overlapping credible intervals around $\mu \approx 25$, suggesting 
no statistically significant separation among them. Lower-ranked systems 
(Base 44 by Wix, Same.new, v0, and Replit) fall below a rating of 24, indicating consistent underperformance against the stronger tools.

These results establish a clear separation between the leading systems and 
the trailing cluster.

\definecolor{gold}{RGB}{255,215,0}
\definecolor{silver}{RGB}{192,192,192}
\definecolor{bronze}{RGB}{205,127,50}

\section{Analysis}\label{sec:analysis}

\subsection{Quality-Gap Analysis}

\subsubsection{Why are scores so different if tools share the same models under the hood?}\label{sec:ai_vs_human}
Several tools likely share foundation models, yet outputs diverge. Lower performers converged on generic templates, with several tools using call-to-action buttons colored in a generic blue repeated across several websites (for instance, see prompt \#7 for v0, and prompt \#17 for Anything in Appendix~\ref{app:generic}), repetitive card grids, minimal interaction, weak responsiveness. Top tools showed deliberate layout planning and asset curation: distinctive typography systems, coherent color programs, higher-quality imagery, and multi-section narratives with clear flow. They also produced working controls, navigation, and subtle animations with fewer broken states. The gap points to differences in prompt handling, template libraries, agent-orchestration techniques, post-processing, and asset pipelines rather than model choice alone.

\subsubsection{Common AI Failure Modes}\label{sec:ai_failures}

We make no claims on AI failure modes related to losses clustering around: typography, spacing, poor hierarchy, technically correct but sterile pages without brand voice, uneven padding, mixed iconography, or naive palette and contrast choices that reduce readability or appeal. Future studies will account for expert labels when judging designs.

\section{Discussion}

\subsection{Limitations}
\paragraph{This study is a static shot.} Experts evaluated websites and webapps generated by AI tools between a set period of dates. This benchmark covers a set of leading commercially available AI tools at the time of testing; future releases with new capabilities remain unassessed. 

\paragraph{Figma Make received shorter prompts.} We note that the prompts used for Figma make were not the exact same prompts as those used for all other tools. Figma's tool has a 500 character maximum input limit, while our prompts were generally over 150+ words. We condensed all prompts via ChatGPT by asking it to "Summarize each of these prompts to 500 characters." We release all sumarized prompts given to Figma make in Appendix~\ref{app:toolprompts}. We included Figma's tool despite this limitation because of its relevance among professional designers and AI developers.

\paragraph{Scope.} \textsc{UI-Bench} isolates \emph{visual} craft, which includes layout, typography, colour, hierarchy, and intentionally ignores UX metrics such as load time, accessibility, or code quality.  We do not include a human-designed baseline in this release; therefore, we make no claims about parity with professional designers. Results should be interpreted as relative rankings among AI tools, not absolute judgments of human-level quality. Additionally, because the evaluator pool is dominated by English-speaking professionals, the results may reflect culturally specific aesthetics. Prompts and interfaces are English-only, potentially under-representing designs optimised for other languages or reading directions.

\subsection{Future Work}
Our findings reflect AI tools available in \emph{2025} and desktop layouts, so they may not generalize to future tool releases, mobile layouts, or print media. Judgements from seasoned professionals reflect industry standards of polish, but expert taste correlates only moderately with lay-user appeal \citep{Ngo2003}. Consumer perceptions can be influenced more by brand familiarity or novelty than by subtle typographic refinements. Next steps include testing AI generated designs against human-made designs to measure effect sizes versus AI, and a failure-mode dataset with a public codebook (typography, balance, color/contrast, detail work, responsiveness, brief adherence, resonance), and expert gold labels.

\section{Conclusion}
In this paper, we introduce \textsc{UI-Bench}, a blinded, expert pairwise benchmark anchored to a client-delivery question for AI text-to-app tools. The protocol uses standardized prompts, a full-view requirement, and TrueSkill-based ranking with uncertainty. Across 4{,}000+ matches, we observe clear separation among tools and identify orchestration choices and potential failure modes that could explain preference gaps. We release prompts and host a public leaderboard to enable further work. Generated sites rated by participants will be released soon.

\section{Acknowledgement}
We thank the expert panel for generously contributing their time and judgment; their participation made large-scale, blinded data collection feasible. We also appreciate the early readers \& researchers whose feedback improved the study and materials.


\bibliographystyle{plainnat}    
\bibliography{refs}           

\appendix
\section{Authoring Prompt}
\label{app:authoring-prompt}

The following is the exact prompt used to generate the synthetic client briefs in \S4.1. 
We release this verbatim for reproducibility.

\begin{tcolorbox}[uibox]

\begin{verbatim}
Goal
Write 30 realistic scenarios ("prompts") that a client might give an 
AI website builder. Produce 6 scenarios per category. In each category, 
write 4 websites and 2 webapps.

Type definitions
Website: read-mostly marketing or informational content. Simple forms 
allowed. No auth. No stored user data.
Webapp: stateful experience with saved data and/or auth. At least 2 
screens (e.g., list/table + create/edit form). Include basic validation 
and empty/loading/error states.

What to deliver (per prompt)
Title — e.g., "Indie Coffee Launch Page"
Type — Website or Webapp
Sector — select from provided list: Tech & SaaS (software, AI tools, 
dev platforms), Finance & Fintech (banking, payments, crypto), Health 
& Wellness (clinics, telehealth, fitness, mental health), Hospitality, 
Food & Travel (restaurants, cafes, hotels, tourism), Retail & 
E-commerce (physical/digital goods, fashion, beauty), Entertainment, 
Sports & Events (artists, venues, tickets, conferences), Education & 
Training (schools, courses, bootcamps), Professional Services (legal, 
accounting, consulting, agencies), Creative & Media (portfolios: 
design, photo, film, content creators), Nonprofit, Community & Public 
Sector (NGOs, churches, clubs, gov), Real Estate & Local Services 
(listings, contractors, trades, salons, repair, pet care, cleaning), 
Industrial, Science & Manufacturing (biotech, hardware, energy), Other 
/ Emerging (catch-all)

Primary goal/CTA: lead / purchase / booking / subscribe / donate / 
apply / RSVP / Informational (no CTA)

Scenario (120–180 words) — a client brief in plain English

Brand vibe — pick 1–2 adjectives and describe it in a sentence 
(e.g., "minimal, editorial" or "bold, playful")

Constraints (optional) — e.g., "Include pricing tiers" or "Include a 
contact form" (no visual or layout directives)

The Five Categories:
Marketing/Landing (conversion-oriented pages, waitlists, SaaS signups)
Editorial/Blog (thought leadership, article hubs, newsletters)
Portfolio/Case Study (agencies, photographers, studios)
E-commerce (small catalogs, digital products, tiered pricing)
Local/Service (restaurants, salons, clinics, trades)

Make sure the scenarios are in the 150+ word range. Double check this 
to confirm. Think carefully as an expert designer.
\end{verbatim}
\end{tcolorbox}

\section{Tool Prompts}
\label{app:toolprompts}

The complete set of 30 prompts is released at: 
\url{https://huggingface.co/datasets/AfterQuery/ui-bench}

Below we include three illustrative examples.

\subsection*{Example 1: Marketing/Landing}
\begin{tcolorbox}[colframe=hotpink!70!black,
                  colback=white,
                  boxrule=1.5pt,
                  arc=3pt,
                  outer arc=3pt,
                  breakable]

\begin{verbatim}
Title: VectorPilot AI — Dev Tool Launch
Type: Website
Sector: Tech & SaaS
Primary goal/CTA: lead
Scenario:
You’re launching VectorPilot, a CLI and dashboard that auto-generates 
API clients from OpenAPI specs and keeps them in sync across repos. 
The page must explain the core value in under 8 seconds: “Ship 
integrations faster. No drift.” Include a hero with a concise promise, 
three outcome-focused benefits, and a 60-second product tour video...
Brand vibe: Minimal and technical, expressed with clean typography, 
disciplined spacing, and terse copy.
Constraints: Include comparison table, FAQ, and a short video embed.
\end{verbatim}
\end{tcolorbox}

\subsection*{Example 2: Editorial/Blog}
\begin{tcolorbox}[colframe=hotpink!70!black,
                  colback=white,
                  boxrule=1.5pt,
                  arc=3pt,
                  outer arc=3pt,
                  breakable]

\begin{verbatim}
Title: The Incident Ledger
Type: Website
Sector: Professional Services
Primary goal/CTA: subscribe
Scenario:
Publish a thought-leadership hub on reliability, incident response, and 
postmortems for CTOs and SRE leaders. The homepage features a concise 
positioning statement and top three pillar articles. Add topic filters, 
author bios, and a “How we write” style guide explaining blameless 
analysis and evidence standards...
Brand vibe: Rigorous and transparent, with sober design and careful 
sourcing as the default.
Constraints: Include editorial policy, glossary, and newsletter sign-up.
\end{verbatim}
\end{tcolorbox}

\subsection*{Example 3: Portfolio/Case Study}
\begin{tcolorbox}[colframe=hotpink!70!black,
                  colback=white,
                  boxrule=1.5pt,
                  arc=3pt,
                  outer arc=3pt,
                  breakable]

\begin{verbatim}
Title: Forge Labs — Product Studio Cases
Type: Website
Sector: Professional Services
Primary goal/CTA: lead
Scenario:
Forge Labs designs and ships MVPs for venture-backed startups. Build a 
portfolio site that opens with a crisp positioning line and a grid of 
selected case studies. Each case has a one-minute read: problem, 
approach, outcome, and a single metric. Include a services summary, 
an engagement model section, and a transparent “How we scope in 2 
weeks.” Add a short contact form that routes to email, no login or 
stored profiles...
Brand vibe: Pragmatic and credible, signaling senior execution without 
theatrics.
Constraints: Case grid, services overview, and short contact form.
\end{verbatim}
\end{tcolorbox}

\section{Generated Sites}
The generated sites viewed by raters will be released soon.

Below we include three illustrative examples of sites viewed by raters.

\subsection*{Example 1: Marketing/Landing}

{\centering
\includegraphics[width=\linewidth]{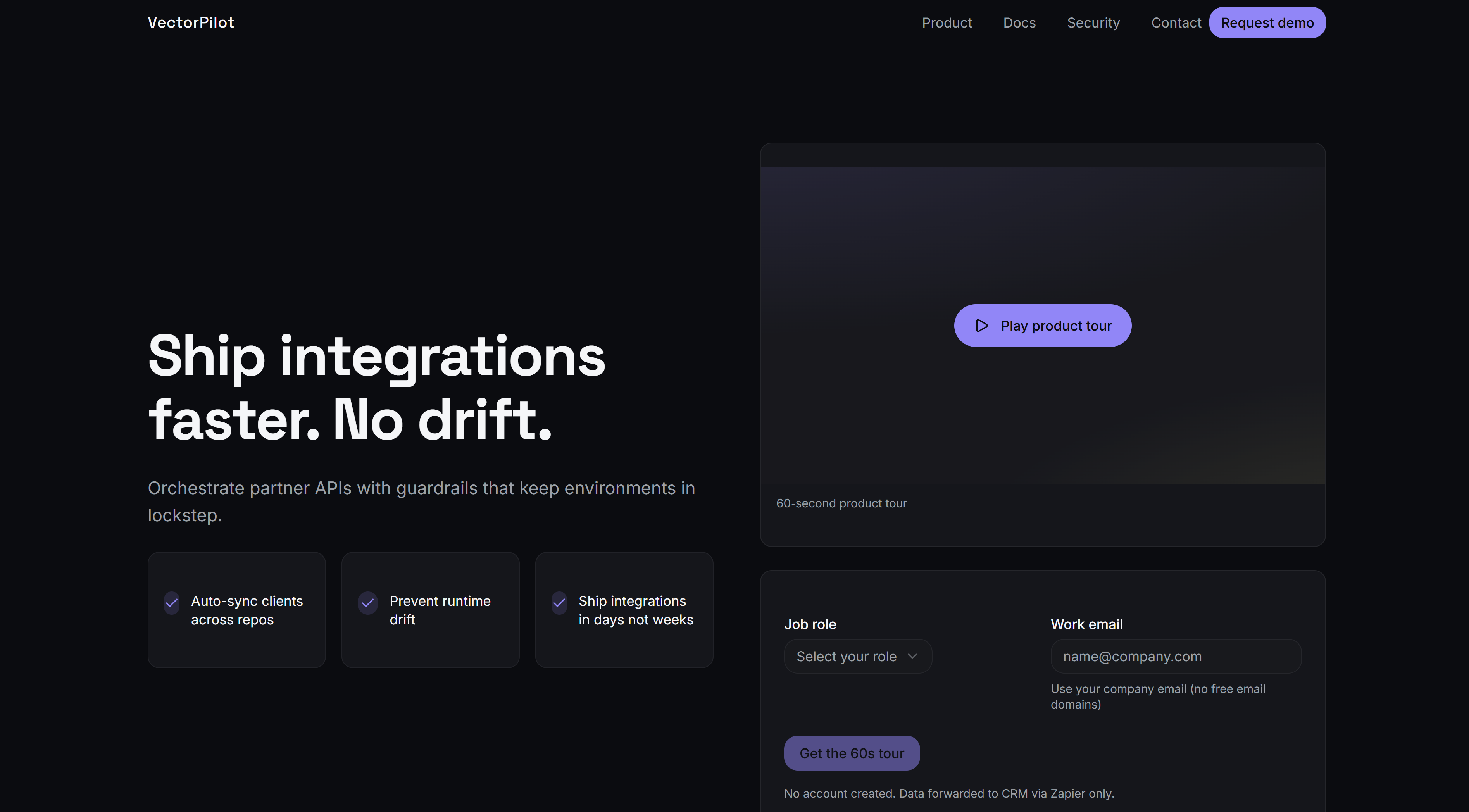}\par
}

\subsection*{Example 2: Editorial/Blog}

{\centering
\includegraphics[width=\linewidth]{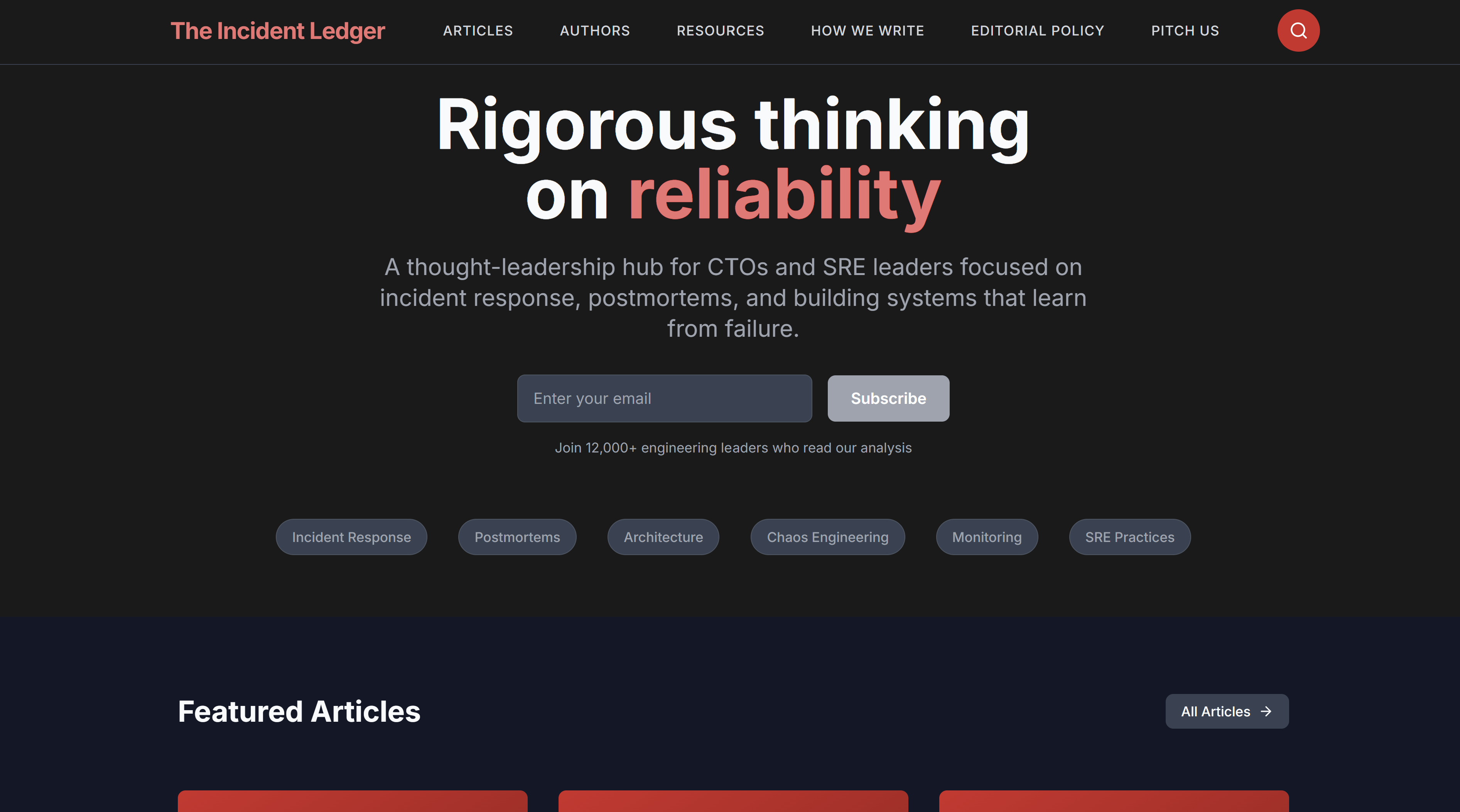}\par
}

\subsection*{Example 3: Portfolio/Case Study}

{\centering
\includegraphics[width=\linewidth]{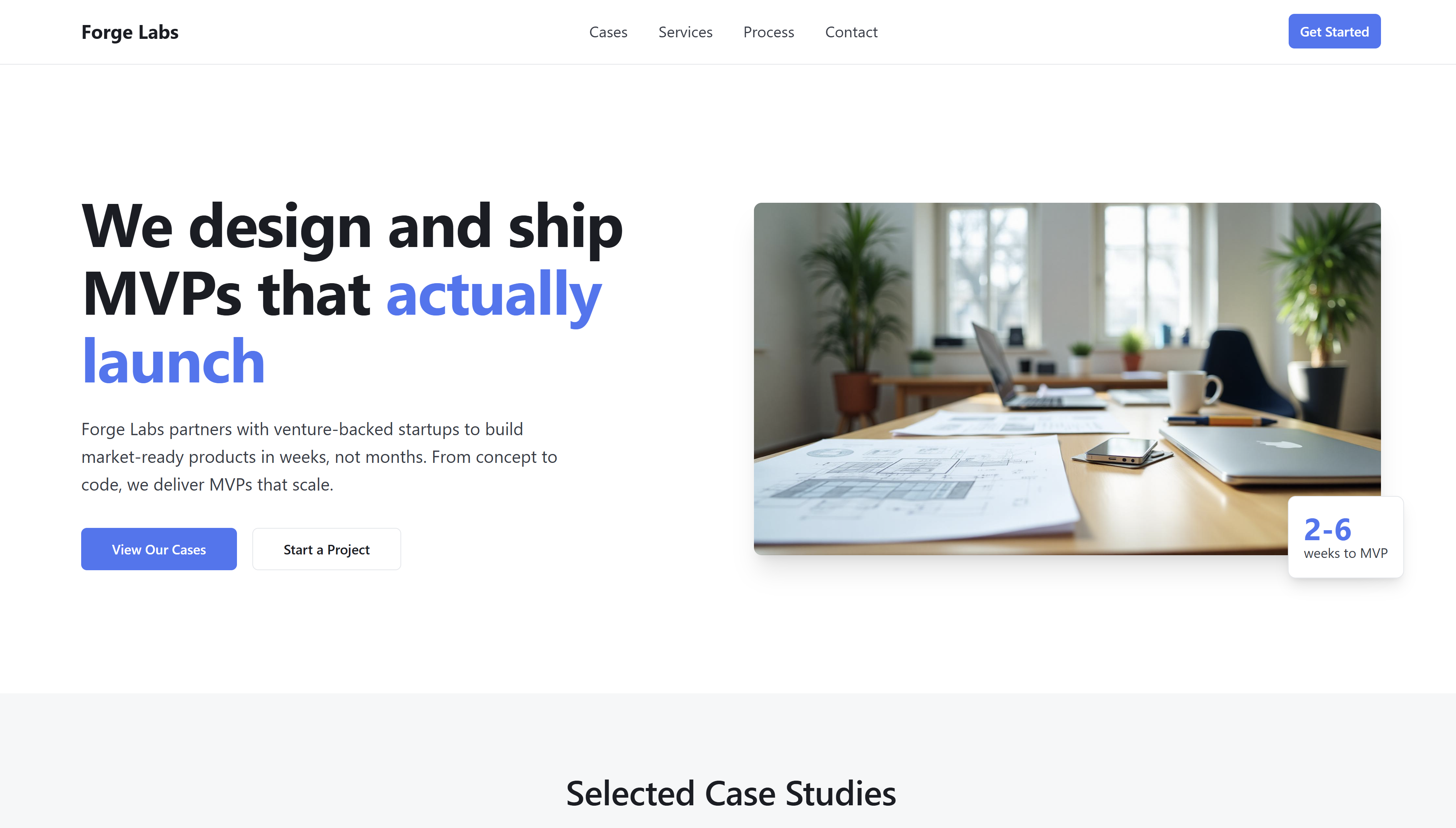}\par
}

\section{Generic Buttons}
\label{app:generic}

\subsection*{Create Anything, Prompt \#17}
{\centering
\includegraphics[width=\linewidth]{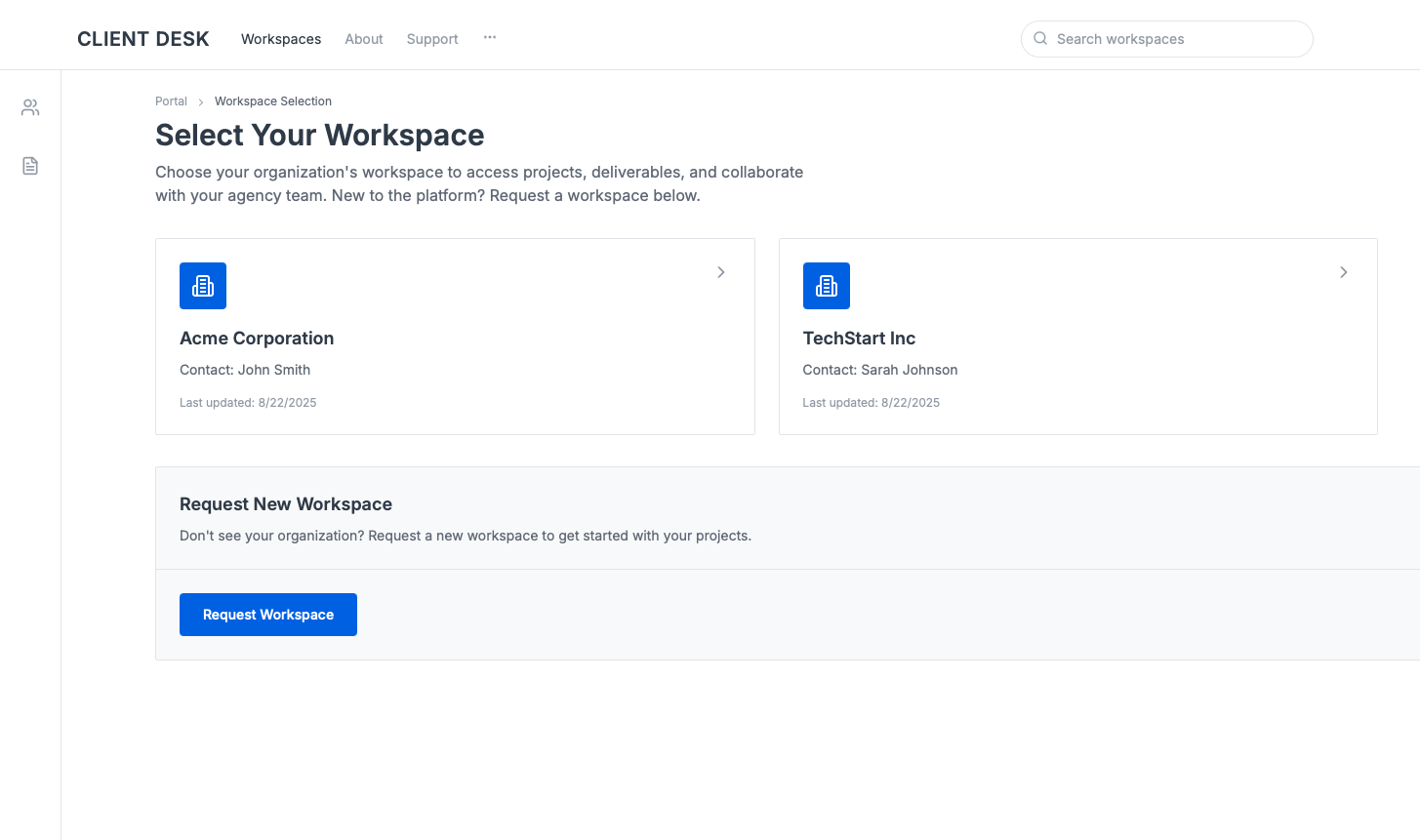}\par
}

\subsection*{v0, Prompt \#7}

{\centering
\includegraphics[width=\linewidth]{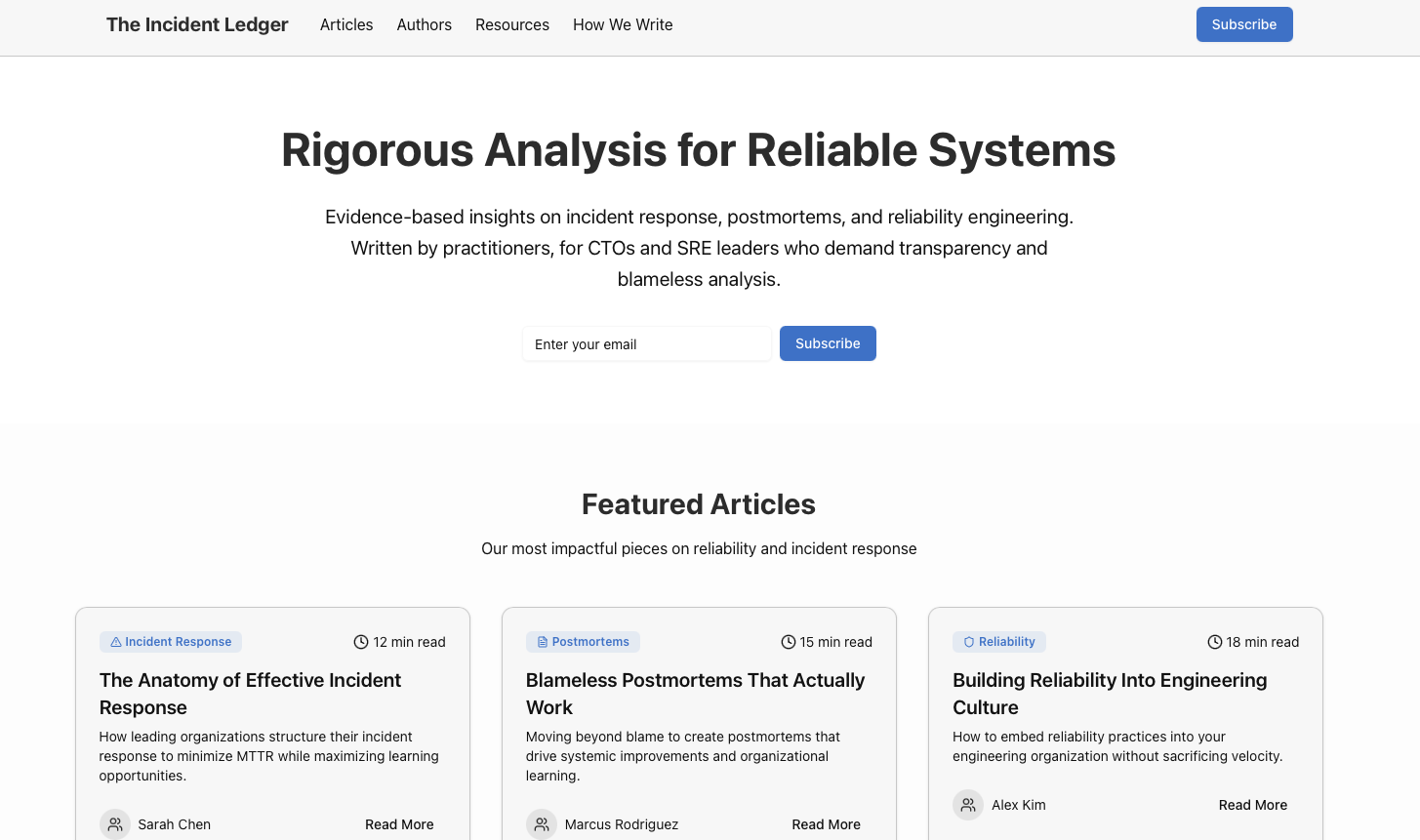}\par
}

\end{document}